\documentclass[10pt, journal, a4paper]{IEEEtran}
\usepackage{lipsum}
\usepackage{mathtools}
\usepackage{cuted}
\usepackage{cite}
\usepackage{cleveref}
\usepackage{multirow}
\usepackage{graphicx}
\usepackage[left=1.6cm, right=1.6cm, top=2.0cm, bottom=2.0cm]{geometry}

\begin{document}
	\title{State-of-the-Art Economic Load Dispatch of Power Systems Using Particle Swarm Optimization}
	\author{Mahamad~Nabab~Alam
		\thanks{The author is with the Department of Electrical Engineering, Indian Institute of Technology, Roorkee, India (e-mail: itsmnalam@gmail.com)}}

	\maketitle
	\begin{abstract}
		Metaheuristic particle swarm optimization (PSO) algorithm has emerged as one of the most promising optimization techniques in solving highly constrained non-linear and non-convex optimization problems in different areas of electrical engineering. Economic operation of the power system is one of the most important areas of electrical engineering where PSO has been used efficiently in solving various issues of practical systems. In this paper, a comprehensive survey of research works in solving various aspects of economic load dispatch (ELD) problems of power system engineering using different types of PSO algorithms is presented. Five important areas of ELD problems have been identified, and the papers published in the general area of ELD using PSO have been classified into these five sections. These five areas are (i) single objective economic load dispatch, (ii) dynamic economic load dispatch, (iii) economic load dispatch with non-conventional sources, (iv) multi-objective environmental/economic dispatch, and (v) economic load dispatch of microgrids. At the end of each category, a table is provided which describes the main features of the papers in brief. The promising future works are given at the conclusion of the review.
	\end{abstract}
	
	
	\IEEEpeerreviewmaketitle
	\section{Introduction}
	\IEEEPARstart{E}{conomic} operation of an electric power system involves unit commitment (UC) and economic load dispatch (ELD). The first one is related to the optimum selection of generating units from available options to supply a particular load demand economically, whereas the second one is related to the optimum power generation from each of the committed (selected) generating units to supply dynamically varying load demand economically \cite{99376}. Proper handling of these two issues not only reduces fuel consumption costs significantly but also reduces transmission losses as well as environmental emission considerably. The issues related to the economic operation of power systems have been widely studied in various books \cite{wood2013power,cata2012electric,lee2008modern,kothari2010power,zhu2015optimization}.
	
	Normally, the ELD problem is formulated as an optimization problem where the objective is to minimize the total cost of fuel consumption while supplying the given load demand successfully and maintaining system operation within the specified limits \cite{kothari2010power}. Commonly, the fuel consumption cost is represented as a simple quadratic function of power generation of the committed generating units along with many non-linear characteristics of that unit. Further, a set of equality and inequality constraints are considered in this minimization problem. Also, some additional non-linear features like valve-point loading effects and multi-fuel input options are considered in the objective function which makes the optimization problem non-convex. Furthermore, generators prohibited operating zones and ramp rate limits make the overall optimization problem exceedingly complex \cite{zhu2015optimization}.
	
	Mathematical techniques like Gradient method, Basepoint and participation factor method, Newton method, and Lambda-iteration method have already been found to be ineffective in solving ELD problems of modern power systems. Also, dynamic programming, non-linear programming, and their modified versions suffer from dimensionality issues in solving ELD problems of modern power systems which are having a large number of generating units. Recently, different metaheuristic optimization approaches have proven to be very effective with promising results in solving ELD problems, such as, simulated annealing (SA) \cite{4631013},  tabu search (TS) \cite{1270429}, artificial neural network (ANN) \cite{387935}, pattern search (PS) \cite{4437387}, evolutionary programming (EP) \cite{485992}, genetic algorithm (GA) \cite{4140683}, differential evolution (DE) \cite{4295007},  and particle swarm optimization (PSO) \cite{7088739,7853592}. Metaheuristic algorithms provide high-quality solutions in relatively less time in solving highly constrained problems \cite{6959090}. Among these algorithms, PSO has shown great potential in solving ELD problems efficiently and effectively \cite{Mahor20092134}. The simple concept, fast computation, and robust search ability are considered to be the most attractive features of PSO.
	
	Although PSO is a very efficient algorithm in solving ELD problems, however, it may suffer from trapping into local minimums during the search process. To handle such trapping into local minima, many modified and hybrid versions of PSO algorithm have been developed for solving ELD problems. Valley et al. \cite{4358769}, have presented PSO, its variants and their applications in solving various issues of power systems in a very comprehensive way. AlRashidi et al. \cite {4358752}, have presented another comprehensive survey considering the application of different PSO algorithms in solving ELD problems.  Lee et al. \cite {4075742}, have discussed the merits and demerits of PSO in solving ELD problems of power system operations. As a large number of publications involving the solution of ELD problems using various PSO algorithms are available in the literature, so a new literature review is needed to obtain a broad idea about the ability of PSO in solving ELD problems in modern power systems prospectives. 
	
	This paper presents a comprehensive survey of the application of PSO in solving ELD problems in electric power systems. Initially, ELD problem formulation and the concept of the PSO algorithm have been discussed. After that, this survey paper covers 14 years of publications from 2003 to 2016 and discusses some of the important contributions available by reputed publishers. The published papers have been classified into five different categories and discussion related to their problem formulation, PSO methodology used, testing of the technique for the formulated model, the output results, and its effectiveness have been analyzed. These five categories are (i) single objective economic dispatch, (ii) dynamic economic load dispatch, (iii) economic dispatch with non-conventional sources, (iv) multi-objective environmental/economic dispatch, and (v) economic dispatch of microgrids.
	
	The remainder of this paper is organized as follows. In Section 2, details of ELD problem formulation is described. In Section 3, the concept of the PSO algorithm is discussed. Review of the application of PSO for solving ELD problems is discussed in Section 4. Finally, conclusions are presented in Section 5.
	
	\section{Review of problem formulation of economic load dispatch problems}
	
	The ELD problem is formulated as an optimization problem of minimization of total cost of power generation to meet a particular load demand subjected to the constraints related to the generator's power output.
	
	\subsection{Conventional problem formulation}
	
	Mathematically, ELD problem can be formulated as \cite{7853592}:
	\begin{equation}
		\label{eq:name1}
		FT=\min\sum_{i=1}^{n} FC_i(PG_i)
	\end{equation}
	
	In eqn. (\ref{eq:name1}), $FT$ is the total cost of generation, $PG_i$ is the output power $i^{th}$ generator, $FC_i(\cdot)$ is the cost function of $i^{th}$ generator and $n$ is the number of generating unit in the system. The cost function of a $i^{th}$ generator is expressed as \cite{4358769};
	\begin{equation}
		\label{eq:name2}
		FC_i(PG_i) = a_i + b_i PG_i + c_i PG_i^2
	\end{equation}
	
	In eqn. (\ref{eq:name2}), $a_i$, $b_i$ and $c_i$ are coefficients of the cost function of $i^{th}$ generator. 
	
	The objective function of ELD problem defined in eqn. (\ref{eq:name1}) is subjected to various constraints which are defined as follows \cite{7853592, 4358769, 4075742};
	
	\subsubsection{Requirement related to power balance}
	
	Mathematically, the requirement related to power balance of any power system area is defined as follows \cite{7853592};
	\begin{equation}
		\label{eq:name3}
		\sum_{i=1}^{n} PG_i =  PD + PN_{loss} +SR
	\end{equation}
	
	In eqn. (\ref{eq:name3}), $PN_ {loss} $ is the total electric power loss in the transmission network, $PD$ is the sum of total power demand of the area and $SR$ is some excess power requirement which is known as spinning reserve. The total electric power loss in the transmission network is defined as follows \cite{7853592};
	\begin{equation}
		\label{eq:name4}
		PN_{loss} = \sum_{i=1}^{n} \sum_{j=1}^{n} PG_i B_{i,j} PG_j + \sum_{j=1}^{n} B_{01} PG_j+B_{00}
	\end{equation}
	
	In eqn. (\ref{eq:name4}), $B_{00}$, $BB_{01}$ and $BB_{i,j}$ are the coefficients of power loss in the transmission networks. 
	
	\subsubsection{Requirements related to power generation capacity of generating unit} 
	
	The power generation from any generator must be within their minimum and maximum generation capacity. Boundary limits on each generator is expressed as follows \cite{7853592,4075742};
	\begin{equation}
		\label{eq:name5}
		\textit{PG}_{i,min} \le \textit{PG}_i \le \textit{PG}_{i,max}
	\end{equation}
	
	In eqn. (\ref{eq:name5}), $PG_{i,min}$ is the minimum power generation and $PG_{i,max}$ is the maximum power generation limits of $i^{th}$ generator. 
	
	\subsubsection{Requirements related to generator ramp rate limits}
	The power output of generating units in certain interval are subjected to the following set of constraints known as ramp rate limits and are expressed as follows \cite{7853592, 4358769, 4075742};
	\begin{eqnarray}
		\label{eq:name5a}
		PG_i-PG_{i,0} \le UR_i \\
		\label{eq:name6b}
		PG_{i,0}-PG_i \le DR_i
	\end{eqnarray}
	\begin{equation}
		\label{eq:name6}
		\max (PG_{i,min}, PG_{i,0}-DR_i) \le PG_i \le \min (P_{i,max},PG_{i,0}+UR_i)
	\end{equation}
	
	In eqn. (\ref{eq:name6}), $UR_i$ is ramp-up, $DR_i$ is ramp-down and $PG_{i, 0}$ is the previous generator output of $i^{th}$ generator.

	\subsubsection{Requirements related to prohibited operating zones of generator}
	
	The smooth power output within the minimum and maximum generation limits of the generators is not possible. There are always some intervals in which power generation are not available. This unavailable power intervals are known as prohibited operating zones of generator and are expressed as follows \cite{7853592, 4075742};
	\begin{eqnarray}
		\label{eq:name7a}
		\textit{PG}_{i,min} \le \textit{PG}_i \le \textit{PG}_{i,1}^{lower} \\
		\label{eq:name7b}
		\textit{PG}_{i,j-1}^{upper} \le \textit{PG}_i \le \textit{PG}_{i,j}^{lower} \\
		\label{eq:name7c}
		\textit{PG}_{i,PZ_i}^{upper} \le \textit{PG}_i \le \textit{PG}_{i,max} 
	\end{eqnarray}
	\begin{equation*}
		\forall j=2,3,...,PZ_i \ \forall i=1,2,...,n_{PZ}
	\end{equation*}
	
	In eqns. (\ref{eq:name7a})-(\ref{eq:name7c}), $PG_{i,j}^{lower}$ is the lower and $PG_{i,j}^{upper}$ is the upper prohibited operating zone for $\forall i=1,2,...,n_{PZ}$ and $\forall j=1,2,...,PZ_i$ for $i^{th}$ generator. Further, $PZ_i$ is the number of the prohibited zones of $i^{th}$ generator and $n_{PZ}$ is the number of generators with such zones.
	
	\subsubsection{Requirements related to the spinning reserve of the area}
	
	Any power generating station must have a certain excess generation to supply during emergency or peak loading conditions. A constraint is imposed to handle the spinning reserve requirement of the system as follows \cite{kothari2010power};
	
	\begin{equation}
		\label{eq:name8}
		\sum_{i=1}^{n} SPG_{i} \ge SR
	\end{equation}
	
	In eqn. (\ref{eq:name8}), $SPG_{i} = \max(PG_{i,max}-PG_i,SPG_{i,max})$ is the reserve contribution of $i^{th}$ generator, $SPG_{i,max}$ is the maximum reserve contribution of $i^{th}$ generator and $SR$ is the surplus spinning reserve capacity after load demand is met. 
	
	\subsection{Some additional cost functions}
	Many cost functions are available in the literature to express practical effects affecting cost functions of generating units committed to supplying a given load demand. In this work, some commonly used cost functions in the literature are considered.
	
	\subsubsection{Cost function with valve-point effects}
	Cost of generation of electricity varies with multiple valve opening and closing. The effects of the valve-point of the cost function is expressed as follows \cite{7853592, 4358769, 4075742}:
	\begin{equation}
		\begin{aligned}
			\label{eq:name10}
			FC_i(PG_i) = a_i + b_i PG_i + c_i PG_i^2+|e_i \times sin(f_i \\ 
			\times (PG_{i,min}-PG_i))|
		\end{aligned}
	\end{equation}
	
\begin{figure}[!h]
		\centering
		\includegraphics[width=0.98\linewidth]{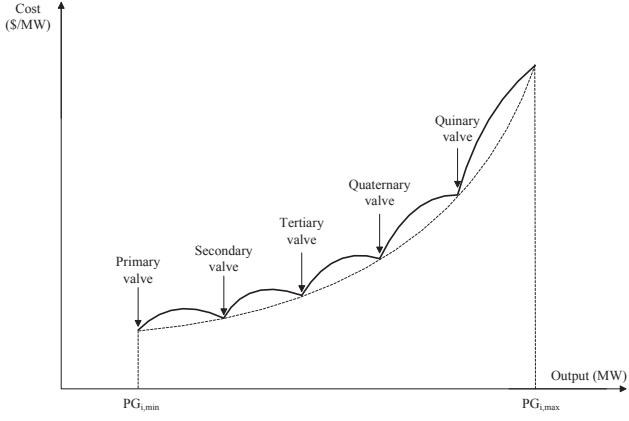}
		\caption{A typical characteristic of cost function of generator with valve-point effects.}
		\label{fig:valve}
	\end{figure}

	In eqn. (\ref{eq:name10}), $e_i$ and $f_i$ are coefficients related to the valve-point effects of $i^{th}$ generator. Fig. \ref{fig:valve} shows the characteristic of the cost function of a generator with the valve-point effects.
	
	\subsubsection{Cost function with multiple fuel input options}
	Modern generators are equipped to use multiple types of fuel options for generating electric power. The cost function with multiple fuel input options of generator is expressed as follows \cite{4358769, 4075742};
	\begin{equation*}
		F_i (PG_i) = \left\{
		\begin{array}{ll}
			a_{i1} + b_{i1} PG_i + c_{i1} PG_i^2, \\ \qquad \qquad \ fuel\ 1, \ PG_{i,min} \le PG_i \le PG_{i,1} \\
			a_{i2} + b_{i2} PG_i + c_{i2} PG_i^2, \\ \qquad \qquad \ fuel\ 2, \ PG_{i,1} \le PG_i \le PG_{i,2} \\      
			... \\
			a_{ik} + b_{ik} PG_i + c_{ik} PG_i^2, \\ \qquad \qquad \ fuel\ k, \ PG_{i,k-1} \le PG_i \le PG_{i,max} \\      
		\end{array}
		\right.
	\end{equation*}
	
\begin{figure}[!ht]
		\centering
		\includegraphics[width=0.98\linewidth]{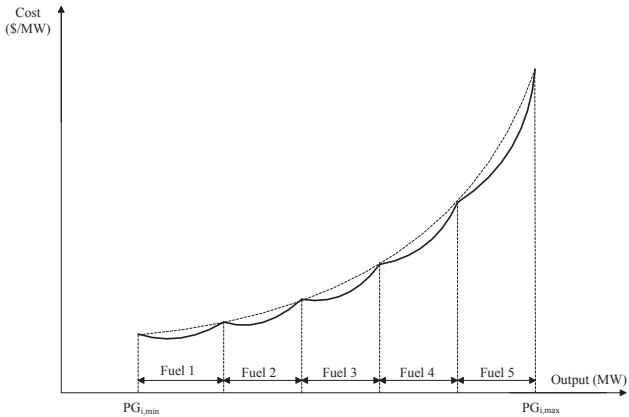}
		\caption{A typical characteristic of cost function of generator with multiple fuel input options.}
		\label{fig:multi}
	\end{figure}

	In the above equations, $a_{ik}$, $b_{ik}$ and $c_{ik}$ are the cost coefficients of $i^{th}$ generator for $k^{th}$ type of fuel. A typical characteristic of cost function with multiple fuel input options is shown in Fig. \ref{fig:multi}.
	
	\subsubsection{Cost function with valve-point effects and multiple fuel input options}
	The electricity generation cost function with valve-point effects and multiple fuel input options is expressed as follows \cite{4358769, 4075742};
	\begin{equation*}
		F_i (PG_i) = \left\{
		\begin{array}{ll}
			a_{i1} + b_{i1} PG_i + c_{i1} PG_i^2 + |e_{i1} \times sin(f_{i1}
			\\ \times (PG_{i,min}-PG_i))|, \\ \qquad \qquad \ fuel\ 1,\ PG_{i,min} \le PG_i \le PG_{i,1} \\
			a_{i2} + b_{i2} PG_i + c_{i2} PG_i^2 + |e_{i2} \times sin(f_{i2} \\ \times (PG_{i,min}-PG_i))|, \\ \qquad \qquad \ fuel\ 2,\ PG_{i,1} \le PG_i \le PG_{i,2} \\      
			... \\
			a_{ik} + b_{ik} PG_i + c_{ik} PG_i^2 + |e_{ik} \times sin(f_{ik} \\ \times (PG_{i,min}-PG_i))|, \\ \qquad \qquad \ fuel\ k,\ PG_{i,k-1} \le PG_i \le PG_{i,max} \\      
		\end{array}
		\right.
	\end{equation*}
	
	In the above eqns., $e_{ik}$ and $f_{ik}$ are the cost coefficients corresponding to valve-point effects and multiple fuel input options of $i^{th}$ generator for $k^{th}$ type of fuel.
	
	\subsubsection{Cost function with emissions of harmful gases in the environment}
	
	Thermal power plants emit harmful gases in the environment constantly \cite{6814462,Roy2013937}. A penalty is imposed by the authorities to limit emission to the minimum level. The cost function representing this effect is expressed as follows;
	\begin{equation}
		\begin{aligned}
			\label{eq:name11}
			E_i(PG_i) = \alpha_i + \beta_i PG_i + \gamma_i PG_i^2+ \zeta_{1i}\times exp(\lambda_1 \times PG_i)+\\
			\zeta_{2i}\times exp(\lambda_2 \times PG_i)
		\end{aligned}
	\end{equation}
	
	In eqn. (\ref{eq:name11}), $\alpha_i$, $\beta_i$, $\gamma_i$, $\zeta_{1i}$, $\zeta_{2i}$, $\lambda_1$ and $\lambda_2$ are the coefficients of emission function.
	
	\subsubsection{Cost function in the presence of wind turbines}
	
	The objective cost function in the presence of wind turbines is normally expressed by the following equation as expressed in \cite{Boqiang20092169};
	
	\begin{equation}
		\label{eq:name12}
		FC_W = \sum_{t=1}^{T} \sum_{i=1}^{nw} fw_i (PG_i^t) S_i^t + C_i S_i^t(1-S_i^{t-1})
	\end{equation}
	
	In eqn. (\ref{eq:name12}), $fw_i(\cdot)$ is the cost function of $i^{th}$ wind turbine,  $S_i^t$ is state of  $i^{th}$ generator having value either 0 or 1 (0 is OFF 1 is ON) at time step $t$, $T$ is the maximum time step, $nw$ is the number of wind turbines and $C_i$ is cold start cost of $i^{th}$ generator.
	
	\subsubsection{Cost function of hydroelectric power generation}
	
	The cost function of hydroelectric power generation can be expressed as follows \cite{kothari2010power};
	
	\begin{equation}
		\label{eq:name13}
		FC_H = \gamma_h \sum_{t=1}^{T} fh_i (PG_i^t)
	\end{equation}
	
	In eqn. (\ref{eq:name13}), $FC_H$ is the total cost of hydroelectric power generation, $fh_i (\cdot)$ is a function representing water discharge during hydroelectric power generation and $\gamma_h$ is a factor which converts the rate of water discharge term in equivalent cost. Hydroelectric power generation is subjected to two main constraints as follows;
	
	a) \textit{Reservoir water head limits}:
	\begin{equation}
		\label{eq:name13a}
		\textit{H}_{min} \le \textit{H} \le \textit{H}_{max}
	\end{equation}
	
	In eqn. (\ref{eq:name13a}), $H_{min}$ is the minimum and $H_{max}$ is the maximum water head $H$ of the reservoir.
	
	b) \textit{Water discharge limits}:
	\begin{equation}
		\label{eq:name13b}
		\textit{Q}_{min} \le \textit{Q} \le \textit{Q}_{max}
	\end{equation}
	
	In eqn. (\ref{eq:name13b}), $Q_{min}$ is the minimum and $Q_{max}$ is the maximum possible water discharge $Q$ of the hydroelectric turbine. 
	
	\section{Review of particle swarm optimization}
	Particle swarm optimization (PSO) is a swarm intelligence based nature inspired meta-heuristic optimization technique developed by James Kennedy and Russell Eberhart \cite{488968} in 1995. It is inspired by the social behavior of birds flocking. The swarm is made of potential solution known as particles. Each particle flies in the search space with the certain velocity and keeps a memory of their best position held so far, and the swarm keeps a memory of the overall best position of the swarm obtained by any particle held during the fly. The next position of each particle in the search space is decided by the present movement, best individual position and the best position of the swarm obtained so far. The present movement of the particle is scaled by a factor called inertial weight $w$ whereas individual best, and the overall best experiences are scaled by acceleration factors $c_1$ and $c_2$ respectively. Also, these experiences are perturbed by multiplying with two randomly generated numbers $r_1$ and $r_2$ between [0,1] respectively. The classical PSO algorithm is represented by two mathematical equation described below. 
	
	Let us assume that the initial population (swarm) of size $N$ and dimension $D$ is denoted as $\bf{X}$ = [$\bf{X}$$_1$,$\bf{X}$$_2$,...,$\bf{X}$$_N$]$^T$, where $'T'$ denotes the transpose operator. Each individual (particle) $\bf{X}$$_i$ $(i=1,2,...,N)$ is given as $\bf{X}$$_i$=$[X_{i,1},X_{i,2},...,X_{i,D}]$. Further, the initial velocity held by the swarm is denoted as $\bf{V}$ = [$\bf{V}$$_1$,$\bf{V}$$_2$,...,$\bf{V}$$_N$]$^T$. Thus, the velocity particle $\bf{X}$$_i$ $(i=1,2,...,N)$ is given as $\bf{V}$$_i$=$[V_{i,1},V_{i,2},...,V_{i,D}]$. Here, the index index $j$ varies from 1 to $D$ and the index $i$ varies from 1 to $N$. The detailed algorithms of various methods are described below for the purpose of completeness \cite{Alam2015}.
	\begin{equation}
		\label{eq:name14}
		{V}_{i,j}^{k+1}=w \times {V}_{i,j}^{k}+c_1 r_1 \times ({Pbest}_{i,j}^{k}-{X}_{i,j}^{k})+c_2 r_2 \times ({Gbest}_{j}^{k}-X_{i,j}^{k})
	\end{equation}
	\begin{equation}
		\label{eq:name15}
		X_{i,j}^{k+1}=X_{i,j}^{k}+V_{i,j}^{k+1}
	\end{equation}
	
\begin{figure}[!ht]
		\centering
		\includegraphics[width=0.98\linewidth]{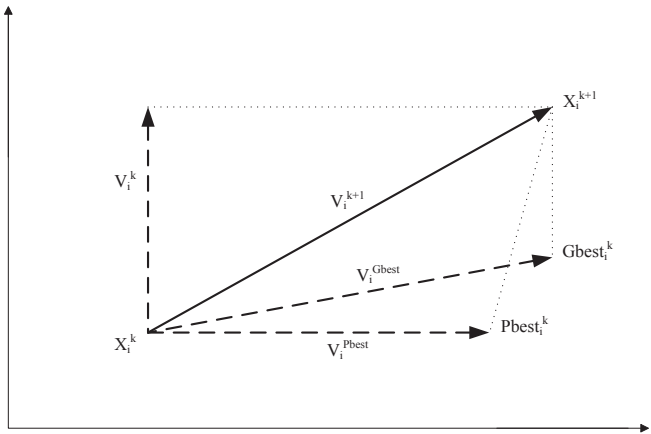}
		\caption{Graphical representation of PSO search mechanism in the search space.}
		\label{fig:pso}
	\end{figure}

	In eqn. (\ref{eq:name14}), $\textit{Gbest}_{j}^k$ represents $j^{th}$ component of the best particle of the population and $\textit{Pbest}_{i,j}^k$ represents personal best $j^{th}$ component of $i^{th}$ particle up to iteration $k$. Fig. \ref{fig:pso} shows the graphical representation of the PSO search mechanism in multidimensional space.
	
	Later, many PSO variants have been developed to improved results. Some of them are discussed below.
	
	\subsection{Time varying inertial weight of PSO}
	Earlier, inertia weight $w$ of the PSO was considered a fixed value between [0.4, 0.9]. Later, it was found that varying inertia with time (iteration) provides faster convergence. Normally, varying inertia with iteration is expressed as follows;
	\begin{equation}
		\label{eq:name16}
		w=w_{max}-k\times(w_{max}-w_{min})/\textit{Maxite}
	\end{equation}
	
	In eqn. (\ref{eq:name16}), $k$ is the current iteration count whereas $Maxite$ is the maximum iteration count set. The value of inertial factor $w$ is used to decrease linearly from $w_ {max} $ to $w_ {min} $ as the iteration increases. 
	
	\subsection{Time varying acceleration factors of PSO}
	Earlier, the values of acceleration factors $c_1$ and $c_2$ of PSO were considered to be equal to 2, but later, it was observed that varying acceleration factors with time (iteration) provide a better solution. Time varying acceleration coefficients (TVAC) of PSO are expressed as follows \cite{6396022};
	\begin{eqnarray}
		\label{eq:name17}
		c_1=c_{1,max}-k\times(c_{1,max}-c_{1,min})/\textit{Maxite} \\
		\label{eq:name18}
		c_2=c_{2,min}+k\times(c_{2,max}-c_{2,min})/\textit{Maxite}
	\end{eqnarray}
	
	In eqn. (\ref{eq:name17}), $c_{1,min}$ is the minimum and $c_{1,max}$ is the maximum limits of acceleration factor $c_1$ whereas in eqn. (\ref{eq:name18}), $c_ {2,min}$ is the minimum and $c_{2,max}$ is the maximum limits of acceleration factor $c_2$. 
	
	\subsection{Constriction factor PSO}
	In \cite{985692}, Maurice Clerc and James Kennedy introduced the concept of constriction factor to PSO to solve problems of multidimensional search space efficiently. PSO with constriction factor shows great potential in solving very complex problems effectively. Mathematically, velocity equation of PSO with constriction factor is represented as follows;
	\begin{equation}
		\label{eq:name19}
		{V}_{i,j}^{k+1}=K[ {V}_{i,j}^{k}+c_1  r_1 \times ({Pbest}_{i,j}^{k}-{X}_{i,j}^{k})+c_2 r_2 \times ({Gbest}_{j}^{k}-X_{i,j}^{k})]
	\end{equation}
	
	In eqn. (\ref{eq:name19}) $K$ is known as the constriction factor of PSO, and is defined as follows;
	\begin{equation}
		\label{eq:name20}
		K=2\kappa/|2-\phi-\sqrt{\phi^2-4\phi}|
	\end{equation}
	where $\phi = c_1+c_2 > 4$ and $\kappa \in [0,1]$ \cite{7117665,7456740}.
	
	\subsubsection{PSO algorithm}
	A typical PSO algorithm is given for completeness as discussed in \cite{Alam2015} as follows 
	
	\begin{enumerate}
		\item Set  $w_{min}$, $w_{max}$, $c_1$ and $c_2$ parameters
		\item Initialize positions $\bf{X}$ and velocities $\bf{V}$ of each particle of the population
		\item Evaluate particles fitness i.e., ${F}$$_i^{k}=f(\bf{X}$$_{i}^k), \forall i$ and find the index $b$ of the best particle
		\item Select $\bf{Pbest}$$_{i}^{k}=\bf{X}$$_{i}^{k}, \forall i$ and $\bf{Gbest}$$^k=\bf{X}$$_{b}^k$
		\item Set iteration count $k=1$
		\item $w=w_{max}-k\times(w_{max}-w_{min})/\textit{Maxite}$
		\item Update velocity and position of particles
		${V}$$_{i,j}^{k+1}=w \times {V}$$_{i,j}^{k}+c_1 r_1 ({Pbest}$$_{i,j}^{k}-{X}$$_{i,j}^{k})+c_2 r_2 ({Gbest}$$_{j}^{k}-X_{i,j}^{k}) ; \; \forall j $ and $ \forall i$
		$X_{i,j}^{k+1}=X_{i,j}^{k}+V_{i,j}^{k+1} ; \; \forall j $ and $ \forall i$
		\item Evaluate the fitness of updated particles i.e., $F_i^{k+1}=f(\bf{X}$$_{i}^{k+1}), \forall i$ and find the index $b1$ of the best particle at this iteration
		\item Update Pbest of each particle of the population $\forall i$
		If $ F_{i}^{k+1}<F_i^k$ then $\bf{Pbest}$$_{i}^{k+1}=\bf{X}$$_{i}^{k+1}$ else $\bf{Pbest}$$_{i}^{k+1}=\bf{Pbest}$$_{i}^k$
		\item Update Gbest of the population
		If $ F_{b1}^{k+1}<F_b^k$ then $\bf{Gbest}$$_{}^{k+1}=\bf{Pbest}$$_{b1}^{k+1}$ and set $b=b1$ else $\bf{Gbest}$$_{}^{k+1}=\bf{Gbest}$$_{}^{k}$
		\item If $k>\textit{Maxite}$ then go to step 12 else $k=k+1$ and go to step 6
		\item Optimum solution is obtained as $\bf{Gbest}$$_{}^{k}$
	\end{enumerate}
	
\begin{figure}[!ht]
		\centering
		\includegraphics[width=0.98\linewidth]{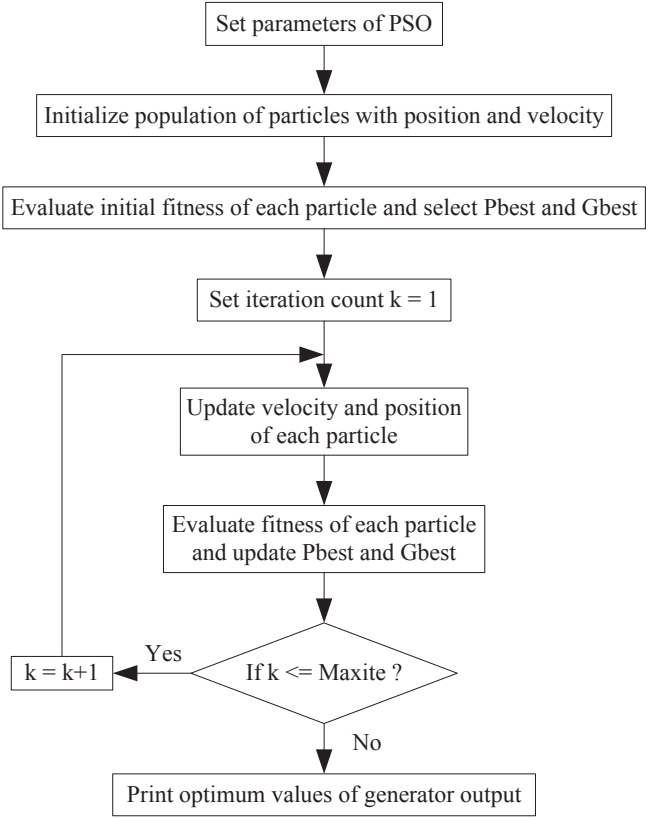}
		\caption{Flowchart of the PSO algorithm.}
		\label{fig:flowc}
	\end{figure}

	Fig. \ref{fig:flowc} shows the flowchart of the PSO algorithm discussed above.
	
	\subsubsection{Parameter selection of PSO}
	Parameter selection of PSO is of extreme importance. Many researchers have given various sets of parameters of the algorithm in the literature. The following parameters of the PSO algorithms are used commonly for solving ELD problems in power systems \cite{699146, Alam2015}:
	
	\begin{itemize}
		\item Population size: 10 to 50
		\item Initial velocity: 10 \% of position
		\item Inertial weight: 0.9 to 0.4
		\item Acceleration factors ($c_1$ and $c_2$): 2 to 2.05
		\item For constriction factors $c_1$ and $c_2$: 2.025 to 2.1  
		\item Maximum iteration (Maxite): 500 to 10000
	\end{itemize}
	
	\section{Review of the application of PSO for solving ELD problems}
	
	The most relevant research papers, in solving ELD problems using PSO, published in years 2002 through 2016, are considered in this article for presentation and discussion. It has been identified five important and related areas of ELD, and the relevant papers published by well-known publishers in the general area of economic dispatch using PSO are classified under one of these five categories. The identified categories are as follows:
	
	\begin{itemize}
		\item Single-objective economic load dispatch (SOELD)
		\item Dynamic economic load dispatch (DELD)
		\item Economic load dispatch with non-conventional sources (ELDNCS)
		\item Multi-objective environmental/economic load dispatch (MOELED)
		\item Economic load dispatch of micro-grids (ELDMG)
	\end{itemize}
	
	At the end of each category, a table is given which gives a brief idea of each of the research papers discussed in detail. 
	
\begin{figure}[!ht]
		\centering
		\includegraphics[width=0.9\linewidth]{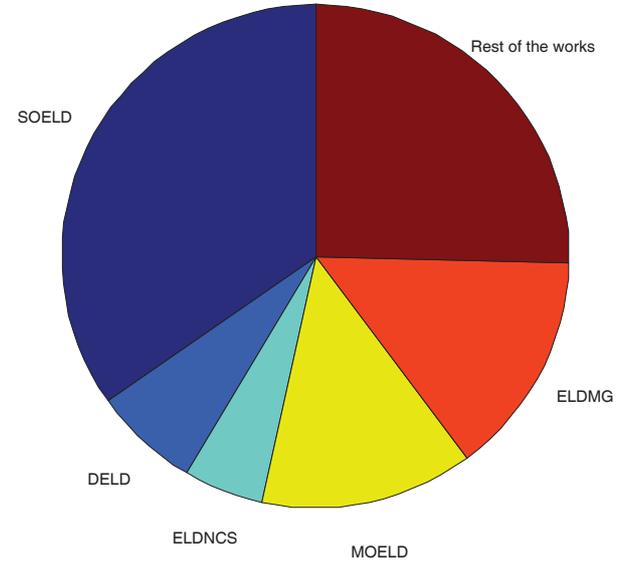}
		\caption{Distribution of the publications in the different areas of economic dispatch using PSO.}
		\label{fig:pie}
	\end{figure}

	Fig. \ref{fig:pie} shows a pie distribution of various publications considered in each of the category discussed above. This pie chart has six division (five for the categories discussed and one for the rest of the discussion including the formulation ELD problem and developed of the classical PSO algorithm). There are 41, 8, 6, 16 and 17 papers under SOELD, DELD, ELDNCS, MOELD, and ELDMG respectively whereas 30 publications are under general discussion about ELD problems and PSO algorithm.
	
\begin{figure}[!ht]
		\centering
		\includegraphics[width=0.98\linewidth]{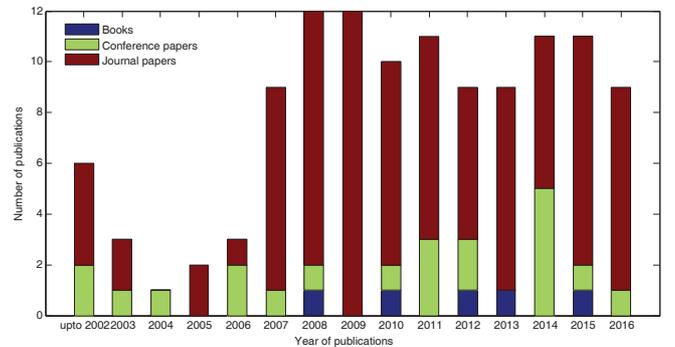}
		\caption{Considered Publications for ELD problems using PSO yearly.}
		\label{fig:bar}
	\end{figure}

	Fig. \ref{fig:bar} shows the year-wise development of all the publications considered in this work. From this figure, it can be observed that there are three types of publication considered which are a) books, b) conference papers and c) journal papers. Most of the paper considered are from various journals.
	
\begin{figure}[!ht]
		\centering
		\includegraphics[width=0.98\linewidth]{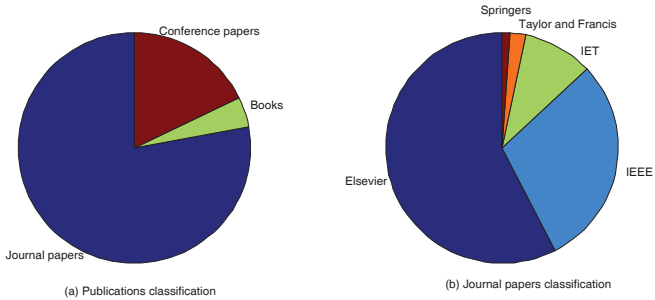}
		\caption{Classifications of all the publications considered in this paper}
		\label{fig:bar2}
	\end{figure}

	Fig. \ref{fig:bar2} shows classification of all the publication considered in this paper. In Fig. \ref{fig:bar2}(a), classification of the entire publications have been shown concerning their type as shown in Fig. \ref{fig:bar} whereas in Fig. \ref{fig:bar2}(b), classification of all the journal papers have been shown as per the database from where the papers have been considered. Under this aegis, 92 journal and 21 conference papers and five books have been considered in this work. Further, out of 92 journal papers, 53 are from $Elsevier$, 27 are from $IEEE$, nine are from $IET$, two are from $Taylor \& Francis$, and one is from $Springer's$ database.
	
	\subsection{Single objective economic load dispatch}
	The single objective economic load dispatch (ELD) deals with only the cost minimization of the generating system while satisfying all the constraints to dispatch power economically. In this section, research papers which possess objective functions of non-smooth, and non-convex, with or without valve-point loading effects, with constraints of generation limits of individual units, multi-fuel input options, ramp rate limits and prohibited operating zones have been analyzed. The problem formulation includes one or more or a combination of two or more practical constraints mentioned above. The different types of PSO algorithms have been used in solving the formulated problems. These PSO algorithms can be categorized in the following four categories:
	
	\begin{itemize}
		\item	ELD using classical PSO
		\item	ELD using a modified/improved/enhanced version of PSO
		\item	Fuzzy controlled PSO
		\item	ELD using a hybrid version of PSO.
	\end{itemize}
	
	Now, research findings of various research papers of each class are explained in detail in the following subsections.
	
	\subsubsection{ELD using classical PSO}
	Park et al. \cite{1270434}, introduced a new approach to the PSO algorithm for solving ELD problems considering non-smooth objective functions. Further, a new constraint handling (CH) approach has been introduced. The suggested CH approach can satisfy the constraints within a reasonable computation time to provide faster convergence. 
	
	Gaing \cite{1216163}, used a basic version of PSO algorithm with a new solution process for solving ELD problems considering generator constraints. The following generator constraints have been considered: a) ramp-rate limits and b) prohibited operating zones. Also, the transmission network losses have also been considered. 
	
	\subsubsection{ELD using modified/improved/enhanced version of PSO}
	
	Park et al. \cite{1388490}, proposed a modified PSO (MPSO) algorithm for solving various types of ELD problems. The cost functions of the ELD problems have been formulated in three ways, a) smooth cost function, b) non-smooth cost function having valve-paint effects and c) non-smooth cost function having multiple fuel input options. In the MPSO algorithm, a strategy has been adopted to enhance the convergence rate and to reduce the search space dynamically. 
	
	Selvakumar and Thanushkodi \cite{4077139}, suggested a new PSO algorithm named NPSO-LRS based on a Local Random Search (LRS) approach for solving the ELD problem. Also, the NPSO-LRS algorithm bifurcates the cognitive behavior into a bad and good experience components. This modification significantly improves the exploration ability of the particles to obtain the global optimum solution. The non-convex ELD problems considered include: a)  valve-point loading effects (EDVL), b) prohibited operating zone (EDPOZ), and c) valve-point loading effects and multiple fuel (EDVLMF) options. 
	
	Cai et al. \cite{Cai2007645}, described a chaotic PSO algorithm (CPSO) by using chaotic local search and an adaptive inertia weight for solving ELD problems considering generator constraints. Also, two CPSO algorithms were presented based on the Tent and logistic equations. 
	
	Selvakumar and Thanushkodi \cite{Selvakumar20082}, proposed a new anti-predatory PSO (APSO) algorithm for solving the EDVL and EDVLMF problems. In PSO, the foraging activity includes the social and cognitive behaviors of the swarm of birds. In the proposed approach, the anti-predatory activity and the foraging activity have been used to help the particles to escape from the predators in the classical PSO. This improves the ability of the particle to escape from local minima and to explore the entire search space efficiently.
	
	Saber et al. \cite{Saber200998}, introduced a modified PSO (MPSO) algorithm for solving ELD problems of a higher order cost function. In the MPSO algorithm, a new velocity vector has been considered. To analyze the importance of the higher order cost function, sensitivity studies of higher order cost polynomials have been performed for ELD problems. 
	
	PSO with crazy particles (PSO-crazy) was used for solving non-convex ED (NCED) problems by Chaturvedi et al. \cite{Chaturvedi2009962}. In the PSO-crazy algorithm, the velocities of crazy particles are randomly adjusted to maintain the momentum of the particles to avoid saturation in the feasible reason. 
	
	Chaturvedi et al. \cite{Chaturvedi2009249}, developed time-varying acceleration coefficients PSO (PSO\_TVAC) algorithm for solving NCED problems of power systems. 
	
	Park et al. \cite{5277440}, introduced an improved PSO (IPSO) algorithm for solving non-convex ELD problems. To escape from a local minimum and to enhance the global search ability of particles, chaotic sequence based inertia-weight and new crossover operation have been proposed in the IPSO algorithm. Further, efficient equality and inequality constraint handling approach has been introduced. 
	
	Meng et al. \cite{5299292}, discussed a quantum-inspired PSO (QPSO) algorithm for solving various types of ELD problems of power systems.  The proposed QPSO employs self-adaptive probability selection and chaotic sequence mutation. 
	
	Zhisheng \cite{Zhisheng20101800}, developed a quantum-behaved PSO, namely QPSO algorithm for solving ELD problems of power systems. 
	
	Neyestani et al. \cite{Neyestani20101121}, introduced a modified PSO (MPSO) algorithm for solving various types of ELD problems. In this algorithm, an attempt has been made to control the diversity of small a population to avoid premature convergence. 
	
	Subbaraj et al. \cite{Subbaraj20101014}, proposed modified stochastic acceleration factors based PSO (PSO-MSAF) algorithm for solving large-scale ELD problems with various generator constraints and transmission losses. 
	
	Safari et al. \cite{Safari20116043}, proposed a new iteration PSO (IPSO) algorithm for solving various types of ELD problems. 
	
	Saber et al. \cite{6039891}, proposed a new hybrid  PSO and DE optimization (PSDEO) algorithm to solve ELD problems of higher order non-smooth cost functions with various practical constraints. The modified PSDEO combines the advantages and disadvantages of both algorithms. In the algorithm, PSO exploits and DE explores the search space to obtain the global optimum solution efficiently. 
	
	Dieu et al. \cite{6039332}, introduced a newly improved PSO (NIPSO) algorithm for solving non-smooth ELD problems. The proposed NIPSO considers time-varying acceleration coefficients (TVAC), a variation of inertia weight with a sigmoid function, particle guidance by pseudo-gradient and quadratic programming (QP) to obtain the initial condition. 
	
	Chalermchaiarbha and Ongsakul \cite{Chalermchaiarbha2013}, proposed a new PSO algorithm called stochastic weight trade-off particle swarm optimization (SWT\_PSO) for solving the ELD problems of power systems. In this algorithm, stochastic inertial weight and acceleration factors are adjusted at each iteration to increase diversity. 
	
	Hosseinnezhad and Babaei \cite{Hosseinnezhad2013}, proposed $\theta$-PSO algorithm for solving various types of ELD problems. In the proposed algorithm, the velocity component of the conventional PSO has been replaced by a phase angle vector. This approach also takes care of various constraints related to transmission loss and generator operational limitations.
	
	Hosseinnezhad et al. \cite{Hosseinnezhad2014}, proposed species based quantum PSO (SQPSO) algorithm for solving different types of ELD problems of power systems. In this SQPSO algorithm, particles are treated as a group at each iteration in addition to the QPSO algorithm. This approach improves the exploration ability of the particles which leads to achieving the global best optimum solution. 
	
	Sun et al. \cite{Sun2014}, discussed random drift PSO (RDPSO) for solving various types of ELD problems of power systems. The concept of RDPSO is inspired by the free electron model of conductors and uses some special evolution equation which leads the particles to reach the optimum global point in the search space. 
	
	Basu \cite{Basu2015}, proposed a modified PSO (MPSO) algorithm for solving different kinds of ELD problems of power systems. In this MPSO, a Gaussian random number generator has been used in calculating the new velocity vector of each particle. This approach improves the exploration ability of the particles and provides much better results.
	
	Hsieh and Su \cite{Hsieh2015}, proposed a new PSO algorithm based on Q-learning for solving the ELD problems. In this algorithm, Q-learning and PSO approaches have been integrated to form the QSO algorithm. In QSO algorithm, the best particle is considered to be the one whose cumulative value of the objective function is the best which is unlike the conventional PSO where the best particle is the one whose current value of the objective function is the best at that particular iteration.
	
	Jadoun et al. \cite{Jadoun2015a}, proposed a modified dynamically controlled PSO (DCPSO) for solving multiple-area ELD problems of power systems. In DCPSO algorithm, velocity vectors of the particles are controlled by introducing exponential constriction functions. This approach helps the particles to explore the entire search space rapidly. In multiple-area ELD modeling, tie-lines have been proposed to exchange power as per the requirement. 
	
	\subsubsection{Fuzzy controlled PSO}
	Niknam \cite{Niknam2010327}, developed a fuzzy adaptive PSO utilizing the Nelder-Mead (FAPSO-NM) algorithm for solving ELD problems. Here at each iteration, the NM algorithm served the objective of a local search for the FAPSO algorithm. Thus, FAPSO-NM improves FAPSO performance significantly. 
	
	Niknam et al. \cite{Niknam20101764}, introduced a novel method, named Fuzzy Adaptive Modified PSO (FAMPSO), for solving non-convex ED problems. A new mutation operator has been introduced to take care of premature convergence. Further, inertia weight and acceleration factors of PSO are tuned using the fuzzy system. 
	
	Mahdad et al. \cite{5771703}, proposed a fuzzy controlled parallel PSO (FCP-PSO) algorithm for solving large-scale non-convex ED problems. In the proposed algorithm, PSO parameters are adjusted dynamically using fuzzy rules. The parallel execution of PSO executed in a decomposed network procedure is found to explore the local search space effectively.
	
	Niknam et al. \cite{Niknam20112805}, described a new adaptive PSO (NAPSO) algorithm suitable for solving ELD problems with multi-fuel input options and prohibited operation zones of generators. Further, a new mutation approach has been used in adaptive PSO (APSO) to escape particles from local minima and search the global optimum. In the APSO algorithm, fuzzy rules have been used to tune the inertial weight of PSO whereas a self-adaptive adjustment approach has been used to adjust other parameters of PSO, such as the cognitive and social parameters. 
	
	Niknam et al. \cite{Niknam20111800}, developed a new hybrid algorithm using variable DE (VDE) and fuzzy adaptive PSO (FAPSO) for solving the ELD problems. The proposed algorithm named FAPSO-VDE considers, the fuzzy rules to adjust parameters adaptively, the PSO to maintain population diversity of the population and the DE to optimize the problem.
	
	\subsubsection{ELD Using Hybrid Version of PSO}
	Chen et al. \cite{4359276}, proposed a new hybrid PSO-RLD algorithm by combining PSO with recombination and dynamic linkage discovery (RLD). The RLD employ a selection operator to adapts the linkage configuration for any type of objective function of the optimization problem. Further, the recombination operator cooperates with PSO by using the configurations which are discovered. 
	
	Coelho and Lee \cite{Coelho2008297}, suggested a combination of chaotic sequences and Gaussian probability distribution functions with PSO in solving ELD problems. The chaotic sequence with logistic map helps the particles to escape from local. The chaotic variables can travel over the whole search space to explore the possibility of the global optimum solution. 
	
	Chaturvedi at el. \cite{4547444}, proposed time-varying acceleration coefficients (TVAC) to PSO to develop a self-organizing hierarchical PSO (SOH\_PSO) algorithm. In this algorithm, the problems of stagnation and premature convergence have been addressed by utilizing reinitialization of velocity vectors and the TVAC respectively. These strategies provide a high-quality, robust solution efficiently even for non-smooth and discontinuous cost functions. 
	
	Kuo \cite{4609943}, proposed a hybrid optimization algorithm using simulated annealing and PSO called SA-PSO. The stochastic search ability of this algorithm pushes the particles to be in the feasible reason for the search space and thus the solution time reduces drastically with a high-quality solution. 
	
	Coelho et al. \cite{Coelho2009510}, proposed a new hybrid chaotic PSO with implicit filtering (HPSO-IF) algorithm for solving ELD problems. In the proposed algorithm, the chaotic sequence provides a high exploration ability in the whole search space, whereas, the fine-tuning of the final results are obtained by the IF in the PSO. 
	
	Vlachogiannis et al. \cite{vlachogiannis2009economic}, introduced a new hybrid optimization algorithm called improved coordinated aggregation-based PSO (ICA-PSO). In this algorithm, all the particles can be attracted by the other particles having better fitnesses in the population than its own except the particle having the best fitness. Further, in this algorithm, the size of the population has been adjusted adaptively. 
	
	Victoire and Jeyakumar \cite{Victoire200451}, presented a hybrid algorithm for solving ELD problems of power systems. The proposed algorithm integrates sequential quadratic programming (SQP) and PSO, named PSO-SQP. In this algorithm, the SQP has been used as fine-tuning of each improved result in the PSO run. 
	
	Sun et al. \cite{Sun20092967}, proposed a modified QPSO with differential mutation (DM) method, named QPSO-DM, for solving ELD problems. 
	
	Kumar et al. \cite{Kumar2011115}, presented a hybrid multi-agent based PSO (HMAPSO) algorithm for solving ELD problems considering valve point effect. The proposed technique integrates a) Multi-agent system (MAS), b) deterministic search, c) PSO and d) bee decision-making process. 
	
	Chakraborty et al. \cite{Chakraborty2011}, proposed a new hybrid PSO algorithm, which is inspired by quantum mechanics for solving various types of ELD problems. The developed algorithm is named hybrid quantum inspired PSO (HQPSO). In HQPSO algorithm, velocity and position vectors of particles are adjusted in a more diverse manner to explore the entire search space to find the global best solution. Also, a special feature has been introduced which increases particle size from single to multiple. 
	
	Sayah and Hamouda \cite{Sayah20131608} proposed two new hybrid methods, a) by combining evolutionary programming (EP) and efficient PSO (EPSO) termed EP-PSO and b) by combining neural network (NN) and efficient PSO (EPSO) termed as NN-EPSO, for solving ELD problems considering valve-point loading effects. 
	
	Abarghooee et al. \cite{6542291}, proposed a hybrid algorithm using an enhanced gradient-based optimization method and a simplified PSO for solving ELD problems of power systems. An attempt has been made to obtain the global or near-global, fast and robust solution in highly constrained ELD problems. 
	
	Table \ref{tab:t1} gives various details like type of algorithm, modeling of ELD problem, the size of the test system, etc.,  for each of the papers reviewed above in this subsection.
	
	\begin{table*}[!ht]
		\renewcommand{\arraystretch}{1}
		\centering
		\caption{Single Objective Economic Load Dispatch}
		\scalebox{0.9}{
			\begin{tabular}{l|l|l|c}
				\hline 
				\textbf{Type of PSO} &	\textbf{Type of the ELD problem} &	\textbf{Test system} &	\textbf{References}	\\
				\hline	\hline	
				PSO with dynamic process	&	SCF, NSCF	&	3-unit	&	\cite{1270434}		\\
				PSO	&	Non-linear with RRL, POZ, NSCF	&	6, 15, 40-unit	&	\cite{1216163}		\\
				MPSO	&	SCF, NSCF with VL effects, NSCF with VL effects and MF input	&	3, 40-unit 	&	\cite{1388490}		\\
				NPSO-LRS	&	Non-convex with EDPO, EDVL, EDVLMF	&	6, 10, 40-unit	&	\cite{4077139}		\\
				Chaotic PSO (CPSO)	&	Non-linear Characteristic of Generators, VL, POZ, RRL, NSCF	&	6, 15-unit	&	\cite{Cai2007645}		\\
				APSO	&	EDVL, EDVLMF	&	10, 40-unit	&	\cite{Selvakumar20082}		\\
				MPSO	&	ELD with Higher Order Cost Function	&	6, 15-unit	&	\cite{Saber200998}		\\
				PSO\_crazy	&	NCED, NCCF	&	3, 6-unit	&	\cite{Chaturvedi2009962}		\\
				PSO\_TVAC	&	NCED, NCCF	&	13, 15, 38-unit	&	\cite{Chaturvedi2009249}		\\
				IPSO 	&	NCCF, POZ, RRL	&	140-unit Korean System	&	\cite{5277440}		\\
				Quantum-inspired PSO (QPSO)	&	NSCF	&	3, 13, 40-unit	&	\cite{5299292}		\\
				Quantum-behaved PSO (QPSO)	&	ELD	&	3, 13-unit	&	\cite{Zhisheng20101800}		\\
				MPSO	&	NSCF with RRLs  and POZs, EDVLMF	&	6-unit	&	\cite{Neyestani20101121}		\\
				PSO-MSAF	&	Large Scale ED Problems with POZ, RRL, Transmission Losses	&	15,40-unit	&	\cite{Subbaraj20101014}		\\
				Iteration PSO (IPSO)	&	Non-continuous, Non-smooth with POZ, RRL	&	6, 15-unit	&	\cite{Safari20116043}		\\
				PSDEO	&	Higher Order NSCF	&	6, 15-unit	&	\cite{6039891}		\\
				NIPSO Based on PSO-TVAC	&	Non-convex	&	13, 40-unit	&	\cite{6039332}		\\
				SWT\_PSO	&	ELD	&	10, 15, 40 &	\cite{Chalermchaiarbha2013}\\
				$\theta$-PSO	&	ELD with generator constraints	&	6, 13, 15, 40 gen. &	\cite{Hosseinnezhad2013}\\
				SQPSO	&	ELD with various other constraints &	6, 15, 40 gen. &	\cite{Hosseinnezhad2014}\\
				RDPSO	&	ELD with generator constraints & 6, 15, 40 gen. &	\cite{Sun2014}\\
				MPSO &	ELD with generator constrants & 3, 6, 40, 140 gen. &	\cite{Basu2015}\\
				QSO & ELD problems & 3, 40 gen. &\cite{Hsieh2015}\\
				DCPSO	&	Multiple area ELD 	&	4, 40, 140 gen.	&	\cite{Jadoun2015a}\\
				FAPSO-NM	&	Non-linear, Non-smooth, and Non-convex with VL Effects	&	13, 40-unit	&	\cite{Niknam2010327}		\\
				FAMPSO	&	Non-convex Economic Dispatch (NCED)	&	13, 40-unit	&	\cite{Niknam20101764}		\\
				FCP-PSO	&	Large Scale Non-convex ED with POZs	&	40-unit	&	\cite{5771703}		\\
				NAPSO, Fuzzy	&	EDLV, EDVLMF and POZs	&	6, 10, 15, 40, 80-unit	&	\cite{Niknam20112805}		\\
				FAPSO-VDE	&	Non-convex ED with Valve-point Loading Effects	&	13, 40-unit	&	\cite{Niknam20111800}		\\
				PSO\_RLD	&	ELD	&	3, 40-unit	&	\cite{4359276}		\\
				PSO, Gaussian and Chaotic Signals	&	Non-linear Generating Characteristics, RRL, POZ	&	15, 20-unit	&	\cite{Coelho2008297}		\\
				SOH\_PSO	&	Non-convex	&	6, 15, 40-unit	&	\cite{4547444}		\\
				SA-PSO	&	Non-linear Cost Function	&	6, 13, 15, 40-unit	&	\cite{4609943}		\\
				HPSO-IF	&	Valve-point Loading Effects	&	13-unit	&	\cite{Coelho2009510}		\\
				ICA-PSO	&	ELD	&	6, 13, 15, 40-unit	&	\cite{vlachogiannis2009economic}		\\
				PSO-SQP	&	Valve-point Loading Effects	&	3, 13, 40-unit	&	\cite{Victoire200451}		\\
				QPSO-DM	&	NSCF, Non-linear Characteristics of Generators, RRL, POZ	&	6, 15-unit	&	\cite{Sun20092967}		\\
				HMAPSO	&	ED with Valve-point Loading Effects	&	13, 40-unit	&	\cite{Kumar2011115}		\\
				HQPSO & ELD with various constraints & 6, 10, 15, 40 gen. system & \cite{Chakraborty2011}\\
				NN-EPSO	&	EDVL	&	13, 40-unit	&	\cite{Sayah20131608}		\\
				EGSSOA	&	POZ, RRL, EDVL, EDVLMF and Transmission Network Losses	&	10, 15, 40, 80-unit	&	 \cite{6542291}		\\
				\hline
				\multicolumn{4}{l}{}\\
				\multicolumn{4}{l}{ED: economic dispatch; VL: valve-point loading; POZ: prohibited operating zones; RRL: ramp rate limits}\\
				\multicolumn{4}{l}{NSCF: non-smooth cost function; NCED: non-convex economic dispatch; NCCF: non-convex cost function}\\
				\multicolumn{4}{l}{NCCF: non-convex cost function; EDVL: ED with valve-point loading; EDVLMF: EDVL and multiple fuels}\\
				\multicolumn{4}{l}{EDVLMF: EDVL and multiple fuels; MF: multiple fuels; MF: multiple fuels; SCF: smooth cost function}\\
				\multicolumn{4}{l}{EDPO: ED with prohibited operation; ELD: economic load dispatch; ELDVL: ELD with valve-point loading.}\\
				
		\end{tabular}}	
		\label{tab:t1}
	\end{table*}	
	
	\subsection{Dynamic economic load dispatch}
	The dynamic economic dispatch problem includes dynamic characteristics or parameters such as spinning reserve constraints, etc., while solving the problem. The dynamic economic dispatch (DED) is the real-time problem in any power system \cite{Zaman2016}. PSO based DED tasks are analyzed to attract the attention of researchers in this particular area.
	
	Victoire et al. \cite{1490578}, proposed a hybrid optimization algorithm considering PSO and SQP for solving reserve constrained DED problems of generator considering valve-point effects. In the proposed hybrid algorithm, the SQP is used as a local optimizer to fine-tune the reason for the solution for the PSO run. Thus, the PSO works as the main optimizer whereas the SQP guides the PSO to obtain better results for solving very DED problems. 
	
	Panigrahi et al. \cite{Panigrahi20081407}, proposed a novel adaptive variable population PSO approach to DED problem of power systems. In the proposed DED model, various system constraints have been considered such as transmission losses, ramp rate limits, prohibited operating zones, etc. 
	
	Baskar and Mohan \cite{Baskar2008609}, suggested an improved PSO (IPSO) algorithm suitable to solve security constraints ELD problems. In the proposed IPSO algorithm constriction factor approach (CFA) has been considered to update the velocity equation of PSO. Security constraints in the paper include bus voltage and line flow limits. 
	
	Baskar and Mohan \cite{Baskar2009615}, proposed an improved PSO (IPSO) algorithm suitable to solve contingency constrained ELD problems. In the proposed IPSO algorithm constriction factor approach (CFA) with eigenvalue analysis has been considered to update the velocity equation of PSO. In the proposed problem formulation, a twin objective a) minimization of severity index and b) minimization of fuel cost have been considered. 
	
	Wang et al. \cite{Wang20102893}, introduced an improved chaotic PSO (ICPSO) algorithm suitable for solving DED problems considering valve-point effects of the generators. In ICPSO algorithm, premature convergence has been controlled using chaotic mutation to improve PSO results. Further, effective constraints handling strategies have been proposed. 
	
	Wang et al. \cite{Wang201114231}, presented a chaotic self-adaptive PSO  (CSAPSO) algorithm suitable to solve dynamic DED problems considering valve-point effects of the generators. In the presented algorithm, an approach has been used to adjust velocity dynamically. Further, a chaotic local search has been used to overcome premature convergence of the algorithm. Also, a random adjustment strategy has been incorporated to handle constraint violations effectively. 
	
	Niknam and Golestaneh \cite{6197805}, proposed an enhanced adaptive PSO (EAPSO) algorithm suitable to solve DED problems. In DED modeling, transmission network losses, ramp-rate limits of generating units and valve-point effects have been considered. In the proposed algorithm, tuning of social and cognitive terns have been proposed to be accomplished dynamically and adaptively. Also, linearly varying inertial weight has been considered. 
	
	Table \ref{tab:t2} gives various details like type of algorithm, modeling of ELD problem, the size of the test system, etc.,  about each paper reviewed above in this subsection.
	
	\begin{table*}[!ht]
		\renewcommand{\arraystretch}{1}
		\centering
		\caption{Dynamic Economic Load Dispatch}
		\scalebox{0.9}{
			\begin{tabular}{l|l|l|c}
				\hline 
				\textbf{Type of PSO} &	\textbf{Type of the ELD problem} &	\textbf{Test system} &	\textbf{References}	\\
				\hline											
				\hline
				Hybrid PSO with SQP	&	Reserve Constrained DED	&	10-unit	&	\cite{1490578}	\\
				Adaptive PSO (APSO)	&	DED with Transmission Losses, RRLs, POZs, NSCF	&	3, 6, 15-unit	&	\cite{Panigrahi20081407}	\\
				IPSO, CFA	&	Security Constraints DED	&	IEEE 14-bus, 66-bus Indian Utility	&	\cite{Baskar2008609}	\\
				IPSO, CFA	&	Contingency Constraints ELD (CCELD) 	&	IEEE 30-bus, IEEE 118-bus	&	\cite{Baskar2009615}	\\
				ICPSO	&	DED with Valve-point Effects	&	10-unit, 30-unit	&	\cite{Wang20102893}	\\
				CSAPSO	&	DED with Valve-point Effects	&	10-unit, 30-unit	&	\cite{Wang201114231}	\\
				EAPSO	&	DED with RRLs and with Valve-point Effects and 	&	10-unit, 30-unit	&	\cite{6197805}	\\
				&Transmission Losses&&\\
				\hline
				\multicolumn{4}{l}{}\\
				\multicolumn{4}{l}{DED: dynamic economic dispatch; TL: transmission line;}\\
				\multicolumn{4}{l}{CCELD: contingency constrained economic load dispatch.}\\
		\end{tabular}}	
		\label{tab:t2}
	\end{table*}	
	
	\subsection{Economic load dispatch with con-conventional sources}
	Non-conventional resources like wind energy, solar energy, tidal energy, etc., considered while solving ELD problems in power systems. These energy sources have almost no fuel cost but may suffer power quality problems. A large number of researchers are working on this issue to make these sources reliable to supply electric energy and synchronized to the grid so that fuel consumption of conventional sources can be reduced considerably \cite{Banerjee2016}. In this section, the research papers are analyzed considering non-conventional sources along with conventional sources, to dispatch electric power economically using various PSO algorithms.
	
	Wang and Singh \cite{4201326}, proposed a modified multi-objective PSO (MPSO) algorithm suitable for solving bi-objective ED problems with wind penetration in the system. Fuzzy rules have been applied to control wind penetration into the system which creates a security problem in the system. A compromise between economic and security requirements has been considered in this paper to achieve both the objectives (economic and security). 
	
	Li and Jiang \cite{5582176}, proposed PSO algorithm-based model to evaluate and lower the risk arises due to high wind power penetration and its variability into the system. In the proposed model, integrated risk management (IRM) and value at risk (VaR) have been used to access the risk and to establish an optimum trade-off between the risk and the profit in the system operations. 
	
	Wu et al. \cite{6038921}, proposed PSO algorithm based ED problems considering combined heat and power (CHP) system consisting thermal, waste heat boiler, gas boiler, fuel cell, PV, wind turbine, battery, and electric load. The random features of PV power, wind power, thermal and electrical load have been handled using chance-constrained programming (CCP). 
	
	Firouzi et al. \cite{BahmaniFirouzi2013232}, proposed a fuzzy self-adaptive learning particle swarm optimization (FSALPSO) algorithm and dynamic economic emission dispatch with wind power and load uncertainties. A roulette wheel technique has been used to model wind power and load uncertainties to generate various scenarios. Also, a fuzzy adaptive approach has been considered to tune algorithm parameters. 
	
	Table \ref{tab:t3} gives various details like type of algorithm, modeling of ELD problem, the size of the test system, etc.,  about each of the papers reviewed above in this subsection.
	
	\begin{table*}[!ht]
		\renewcommand{\arraystretch}{1}
		\centering
		\caption{Economic Load Dispatch with Non-Conventional Sources}
		\scalebox{0.9}{
			\begin{tabular}{l|l|l|c}
				\hline 
				\textbf{Type of PSO} &	\textbf{Type of the ELD problem} &	\textbf{Test system} &	\textbf{References}	\\
				\hline \hline		
				MPSO, Fuzzy	&	ED and Security Impact (Wind and Thermal) 	&	IEEE 30-bus 6-generator	&	\cite{4201326}	\\
				MOPSO, Fuzzy	&	ELD and Security Impact (Wind and Thermal)	&	IEEE 30-bus 6-generator	&	\cite{Wang20081361}	\\
				PSO, CCP	&	Combined Heat and Power Dispatch of micro-grid	&	A typical 6-generator System	&	\cite{6038921}	\\
				PSO	&	Lowering the risk (with Wind Power)  	&	IEEE 30-bus 6-generator, Shanghai Network	&	\cite{5582176}	\\
				FSALPSO, PSO Variants 	&	DEED in presence of wind generation	&	A typical 10-generator system	&	\cite{BahmaniFirouzi2013232} 	\\	
				\hline
		\end{tabular}}	
		\label{tab:t3}
	\end{table*}
	
	\subsection{Multi-objective environmental/economic load dispatch}
	Multi-objective environmental/economic dispatch (EED) problems include not only fuel consumption but also emission of SOx, NOx, etc. while solving economic load dispatch. The EED is a multi-objective problem with objectives of minimizing the emissions as well as the cost of generation \cite{Gholami2014}. There has been a diversity of research about EED problems. Here, it has been analyzed research works dealing with EED problems using various PSO algorithms.
	
	Jeyakumar et al. \cite{Jeyakumar200636}, described the multi-objective problem by means of four different models, namely, a) multi-area ED as MAED, b) piecewise quadratic cost function considering multi-fuel option as PQCF, c) cost as well as emission minimization as CEED and d) ED with the prohibited operating zone as ED with POZ. 
	
	Huang and Wang \cite{4077115}, proposed a novel hybrid optimization technique to considers the network of radial basis function (RBF) for real-time power dispatch (RTPD) by combining enhanced PSO (EPSO) algorithms and orthogonal least-square (OLS) method. The OLS algorithm gives the number of centers in the hidden layer, and the EPSO algorithm gives fine-tuned parameters in the network. 
	
	Wang and Singh \cite{Wang20071654}, proposed fuzzified multi-objective PSO (FMOPSO) algorithm suitable to solve multi-objective EED problems of power systems. Here, minimization of fuel cost and emissions has been considered as objectives to meet. 
	
	AlRashidi and El-Hawary \cite{4349053}, presented a new hybrid optimization algorithm by combining Newton-Raphson and PSO suitable to solve multi-objective EED problems of power system operation. Further, a new inequality constraint handling mechanism has been incorporated into the proposed optimization approach. 
	
	Wang and Singh \cite{Wang20081466}, proposed an improved PSO algorithm considering a combined deterministic and stochastic model of ELD problems and simultaneously considering environmental impact. 
	
	Agrawal et al. \cite{4454712}, proposed a fuzzy clustering-based PSO (FCPSO) algorithm for solving highly constrained multi-objective EED problem involving conflicting objectives. The niching and fuzzy clustering technique has been used to direct the particles towards lesser-explored regions of the Pareto front. Further, an adaptive mutation operator has been used to prevent premature convergence. Also, a fuzzy-based approach has been used to make a compromise with objectives. 
	
	Wang and Singh \cite{Wang2009298}, discussed the solution of multi-area environment/economic dispatch (MAEED) problems using an improved multi-objective PSO (MOPSO). The objectives are to obtain optimum ELD and to minimize pollutant emissions. In the proposed model, tie-line transfer limits and a reserve-sharing scheme have been used to ensure the ability of each area to fulfill its reserve requirement.
	
	Cai et al. \cite{Cai20091318}, introduced a multi-objective chaotic PSO (MOCPSO) algorithm suitable for solving EED problems. The comparison of performances of the proposed MOCPSO and the conventional MOPSO algorithm has been performed. 
	
	Zhang et al. \cite{Zhang2012213}, proposed a bare-bones multi-objective PSO (BB-MOPSO) algorithm suitable for solving EED problems of power systems operations. In this algorithm, constraint handling strategy, mutation operator, crowding distance, fuzzy membership functions and an external repository of elite particles have been used to make BB-MOPSO algorithm much more efficient for multiple objectives optimization problems. 
	
	Chalermchaiarbha and Ongsakul \cite{Chalermchaiarbha2012}, proposed a new elitist multi-objective PSO (EMPSO) algorithm for solving multiple objectives ELD problems of power systems. In the proposed algorithm, fuzzy multi-attribute decision making is utilized to obtain a good compromise among the conflicting objectives. 
	
	Zeng and Sun \cite{Zeng2014}, proposed an improved PSO algorithm for solving the CHP-DED problems with various systems constraints of power systems. In the proposed algorithm, chaotic mechanism, TVAC, and self-adaptive mutation scheme have been considered. Also, various constraints handing approaches have been utilized. 
	
	Jadoun et al. \cite{Jadoun2015}, proposed a new modified modulated PSO (MPSO) algorithm for solving various types of economic emission dispatch (EED) problems of power systems. In this algorithm, the velocity vector of the conventional PSO has been modified by the truncated sinusoidal constriction function in the velocity equation. Further, the conflicting objectives of the EED problem, which is compromised of economic dispatch and emission minimization are combined in a fuzzy framework by suggesting adjusted fuzzy membership functions which are then optimized using the proposed MPSO. 
	
	Jiang et al. \cite{Jiang2015}, proposed a newly modified gravitational acceleration enhanced PSO (GAEPSO) algorithm for solving multiple objectives wind-thermal economic and emission dispatch problems of the power system. In this algorithm, the velocity of each particle is simultaneously updated using PSO and gravitational search algorithm (GSA). The concepts of updating the velocity vector using PSO provides enough exploration whereas the GSA provides enough exploitation to each particle. These features of the proposed GAEPSO make it a faster and more efficient algorithm in solving ELD problems.
	
	Mandal et al. \cite{Mandal2015}, proposed a newly modified self-adaptive PSO algorithm for solving emission constrained economic dispatch problems of power systems. The proposed algorithm is a self-organizing hierarchical PSO with time-varying acceleration coefficients (SOHPSO\_TVAC).  
	
	Liu et al. \cite{Liu2016}, proposed cultural multi-objective QPSO (CMOQPSO) algorithm for solving the EED problem of power systems. In this algorithm, population diversity is maintained by introducing a cultural evolution mechanism in the QPSO algorithm. Believe space, available in the cultural evolution mechanism is utilized to avoid premature convergence. This feature leads to explore the entire search space effectively and gives much better results. 
	
	Table \ref{tab:t4} provides various details, such as the type of algorithm used, modeling of the ELD problem, size of the test system, etc.,  about each of the papers reviewed above in this subsection.
	
	\begin{table*}[!ht]
		\renewcommand{\arraystretch}{1}
		\centering
		\caption{Multi-Objective Environmental/Economic Dispatch}
		\scalebox{0.9}{
			\begin{tabular}{l|l|l|c}
				\hline
				\textbf{Type of PSO} &	\textbf{Type of the ELD problem} &	\textbf{Test system} &	\textbf{References}	\\
				\hline		\hline
				PSO, CEP	&	Multi-fuel Option, Combined EED with RRL	&	6, 10, 15, 16-gen. systems	&	\cite{Jeyakumar200636}	\\
				EPSO, OLS	&	Real-time Power Dispatch	&	IEEE 30-bus six-gen.	&	\cite{4077115}	\\
				FMOPSO	&	Bi-objective Cost as well as Pollutant Emission Minimization	&	14-generator system	&	\cite{Wang20071654}	\\
				PSO, Newton-Raphson	&	Minimization of Real Power Loss, Fuel Cost, and Gaseous emission	&	IEEE 30-bus six-gen.	&	\cite{4349053}	\\
				Improved PSO	&	Deterministic and Stochastic Model of ELD with Environmental Impact	&	IEEE 30-bus six-gen.	&	\cite{Wang20081466}	\\
				FCPSO, Niching	&	Highly Constrained Multi-objective EED	&	IEEE 30-bus six-gen.	&	\cite{4454712}	\\
				MOPSO	&	Reserve-constrained Multi-are EED (MAEED)	&	Typical 7-gen. system	&	\cite{Wang2009298}	\\
				MOCPSO	&	Fuel Cost and Emission Minimization	&	IEEE 30-bus six-gen.	&	\cite{Cai20091318}	\\
				BB-MOPSO, Fuzzy	&	Multi-objective EED	&	IEEE 30-bus six-gen.	&	\cite{Zhang2012213}	\\
				EM PSO & Multi-objective ELD problems with various systems constraints & 6 and 18 gen. & \cite{Chalermchaiarbha2012}\\
				Improved PSO & CHP based DED problems with various systems constraints & 10 gen. & \cite{Zeng2014}\\
				MPSO	&	EED	&	6, 10 and 40-gen. systems &	\cite{Jadoun2015}\\
				GAEPSO	&	wind-thermal economic and emission dispatch	&	6 and 40-gen. systems &	\cite{Jiang2015}\\
				SOHPSO\_TVAC	&	emission constrained EED	& two typical test systems &	\cite{Mandal2015}\\
				CMQPSO	&	EED	&	6 and 40-gen. systems &	\cite{Liu2016}\\
				\hline
		\end{tabular}}
		\label{tab:t4}
	\end{table*}
	
	\subsection{Economic load dispatch of micro-grids}
	The sustainable development goal of countries can be achieved through a provision of access to clean, secure, reliable and affordable energy. This can only be achieved by renewable power generation. To access such electric power we need excellent micro-grids technologies. Various researchers are now focusing on the technical and economical suitability of micro-grids \cite{5743067,6175747,Khare2013,Soares2016,Abdi2016}.
	
	Moghaddama et al. \cite{Moghaddam20121268}, presented a comprehensive literature review on ELD problems related to micro-grids. In this work, the primary focus has been given to the application of PSO algorithms in solving issues related to the economic operations of micro-grids. Basu et al. \cite{6006566}, proposed a CHP-based micro-grids economic scheduling considering network losses. Nikmehr and Ravadanegh \cite{7151035}, proposed the optimum power dispatch of micro-grids considering probabilistic model using PSO. 
	Also, \cite{7042324}, introduced the economic scheduling of multi-micro-grids using PSO. Wu et al. \cite{Wu2014336}, proposed the economic operation of CHP based micro-grid system considering photovoltaic arrays (PV), wind turbines (WT), diesel engines (DE), fuel cells (FC), micro-turbines (MT) and battery system (BS). The proposed model has been solved using an improved PSO. 
	
	Yao et al. \cite{Yao2012}, proposed a quantum-inspired PSO (QPSO) algorithm for solving the green energy based ELD problems of smart grids. In the proposed QPSO, a quantum-inspired evolutionary algorithm (QEA) which is based on quantum computing has been utilized to obtain better and faster global optimum results. In the ELD problem, wind power uncertainty and carbon tax have been considered while formulating the problem. 
	
	Faria et al. \cite{Faria2013}, discussed a modified PSO for solving the ELD problems including the demand response and distributed generation (DG) resources of modern smart grids systems. 
	Hu et al. \cite{Hu2014}, discussed fuzzy-adaptive PSO (FAPSO) algorithm for minimizing distribution network loss using optimum load response to the consumers.  
	
	Wu et al. \cite{Wu2015}, proposed multi-objective PSO (MPSO) for solving the ELD problems of the risk-based wind-integrated power system. The proposed modeling considers wind power uncertainty and optimum power dispatch simultaneously. A probabilistic model has been considered to predict wind availability and risk related to supply the load demand. 
	
	Cheng et al. \cite{Cheng2016}, proposed the PSO algorithm for solving energy management of a hybrid generation system (HGS). The proposed HGS consists of power generation from the photovoltaic array, wind turbine, micro-turbine, battery banks, and the utility grid.   
	
	Elsied et al. \cite{Elsied2016}, proposed energy management of microgrids considering minimization of the energy cost, carbon dioxide, and pollutant emissions while maximizing the power of the available renewable energy resources using binary PSO (BPSO) algorithm. 
	
	Li et al. \cite{Li2016}, proposed chaotic binary PSO (CBPSO) for solving the ELD problems of micro-grids. A new fuzzy-based modeling has been developed to minimize systems losses, pollutant emission and the cost of supplying the power demand. 
	
	Table \ref{tab:t5} gives various details such as the type of algorithm used, modeling of the ELD problem, size of the test system, etc.,  about each of the papers reviewed above in this subsection.
	
	\begin{table*}[!ht]
		\renewcommand{\arraystretch}{1}
		\centering
		\caption{Economic Dispatch in Micrgrids}
		\scalebox{0.9}{
			\begin{tabular}{l|l|l|c}
				\hline 
				\textbf{Type of PSO} &	\textbf{Type of the ELD problem} &	\textbf{Test system} &	\textbf{References}	\\
				\hline	\hline		
				FSAPSO	&	A survey paper in micro-grids power dispatch	& A typical micro-grid test system	&	\cite{Moghaddam20121268}	\\
				PSO	&	CHP-dispatch with losses and emission in micro-grids	&	IEEE 14-bus five-gen.	&	\cite{6006566}	\\
				PSO, ICA	&	Micro-grids with WT and PV power dispatch	&	A typical three-gen. micro-grid test system	&	\cite{7151035}	\\
				PSO	&	Micro-grids with WT and PV with losses power dispatch &	A typical three-gen. micro-grid test system	&	\cite{7042324}	\\
				PSO	&	Micro-grids with PV, WT, DE, FC, MT, BT with losses 	&	A typical seven-gen. micro-grid test system 	&	\cite{Wu2014336}	\\
				&using actual mathematical modelling of the resources &&\\
				QPSO	&	ELD of smart grids with uncertainty and carbon tax	&	Modified IEEE 30-bus six-gen.	&	\cite{Yao2012}	\\
				Modeified PSO	&	ELD with demand response and presence of DG	&	A real test system 	&	\cite{Faria2013}	\\
				FAPSO	&	ELD with demand response from consumers	&	A typical 18-bus system with 3-WT	&	\cite{Hu2014}	\\
				MPSO	&	ELD with risk based WT	&	IEEE 30-bus six-gen.	&	\cite{Wu2015}	\\			
				PSO	&	ELD of HGS with PV, WT, MT, BT and utility system &	IEEE 30-bus six-gen.	&	\cite{Cheng2016}	\\
				BPSO& EED of micro-grids with various renewable resources & A typical micro-grid system &\cite{Elsied2016}\\
				CBPSO, fuzzy	&	Multi objective EED 	&	A typical distribution network	&	\cite{Li2016}	\\			
				\hline
				\multicolumn{4}{l}{}\\
				\multicolumn{4}{l}{CHP: Combined heat and power;}\\
				\multicolumn{4}{l}{HGS: Hybrid generation system;}\\			
				\multicolumn{4}{l}{ICA: Imperialist competitive algorithm;}\\
				\multicolumn{4}{l}{FSAPSO: Fuzzy self adaptive particle swarm optimization;}\\
		\end{tabular}}
		\label{tab:t5}
	\end{table*}
	
	Table \ref{tab:t6} gives more details about the various database and their journals from which the papers have been considered in this study. Further, the serial numbers of all those references taken from a particular journal have also been included in this table. 
	
	\begin{table*}[!ht]
		\renewcommand{\arraystretch}{1}
		\centering
		\caption{Various database and the journals considered in this study}
		\scalebox{0.9}{
			\begin{tabular}{l|l|l|c}
				\hline 
				\textbf{Publishers} &	\textbf{Name of journals} &	\textbf{References} &	\textbf{No. of papers}	\\
				\hline \hline
				\multirow{12}{*}{Elsevier}	&	Applied Energy	&	\cite{Niknam2010327,Soares2016}	&	2	\\	
				&	Applied Soft Computing	&	\cite{Chaturvedi2009962,Niknam20112805,Sayah20131608,Mandal2015,Liu2016,Khare2013}	&	6	\\	
				&	Chaos, Solitons \& Fractals	&	\cite{Coelho2009510}	&	1	\\	
				&	Electric Power Systems Research	&	\cite{Alam2015,Selvakumar20082,Saber200998,Victoire200451,Baskar2009615,Wang20081361,Wang20071654,Wang20081466}	&	8	\\	
				&	Energy	&	\cite{Niknam20101764,BahmaniFirouzi2013232}	&	2	\\	
				&	Energy Conversion and Management	&	\cite{Cai2007645,Chalermchaiarbha2013,Niknam20111800,Sun20092967,Panigrahi20081407,Wang20102893,Cai20091318,Elsied2016}	&	8	\\	
				&	Engineering Applications of Artificial Intelligence	&	\cite{Neyestani20101121,Zaman2016,Wang2009298}	&	3	\\	
				&	Expert Systems with Applications	&	\cite{Zhisheng20101800,Safari20116043,Wang201114231}	&	3	\\	
				&	Information Sciences	&	\cite{Zhang2012213}	&	1	\\	
				&	International Journal of Electrical Power \& Energy Systems	&	\cite{Roy2013937,Chaturvedi2009249,Subbaraj20101014,Hosseinnezhad2013,Hosseinnezhad2014,Basu2015,Jadoun2015a,Coelho2008297,Kumar2011115}	&	15	\\	
				&&\cite{Baskar2008609,Banerjee2016,Jeyakumar200636,Jadoun2015,Jiang2015,Wu2014336}& \\ 
				&	Renewable and Sustainable Energy Reviews	&	\cite{Mahor20092134,Boqiang20092169,Moghaddam20121268}	&	3	\\	
				&	Renewable Energy	&	\cite{Cheng2016}	&	1	\\	\cline{1-4}
				\multirow{5}{*}{IEEE}	&	IEEE Transactions on Evolutionary Computation	&	\cite{4358769,4358752,985692,4454712}	&	4	\\	
				&	IEEE Transactions on Industrial Informatics	&	\cite{Sun2014,Yao2012}	&	2	\\	
				&	IEEE Transactions on Power Systems	&	\cite{99376,387935,485992,1216163,1388490,4077139,5277440,5299292,4547444,4609943}	&	16	\\	
				&&\cite{vlachogiannis2009economic,1490578,5582176,4077115,4349053,6006566}& \\ 
				&	IEEE Transactions on Smart Grid	&	\cite{Abdi2016,7042324,Faria2013,Li2016}	&	4	\\	
				&	IEEE Transactions on Systems, Man, and Cybernetics, Part B: Cybernetics	&	\cite{4359276}	&	1	\\	\cline{1-4}
				\multirow{2}{*}{IET}	&	IET Generation, Transmission \& Distribution	&	\cite{4140683,4295007,Chakraborty2011,6542291,6197805,Gholami2014,Hu2014,Wu2015}	&	8	\\	
				&	IET Renewable Power Generation	&	\cite{5743067}	&	1	\\	\cline{1-4}
				Taylor \& Francis	&	Electric Power Components and Systems	&	\cite{Chalermchaiarbha2012,Zeng2014}	&	2	\\	\cline{1-4}
				Springer's	&	Neural Computing and Applications	&	\cite{Hsieh2015}	&	1	\\	\cline{1-4}
		\end{tabular}}
		\label{tab:t6}
	\end{table*}
	
	\section{Conclusion}
	This paper presents a literature review of the application of particle swarm optimization (PSO) algorithm for solving various types of economic load dispatch (ELD) problems in power systems. A survey of papers and reports of the period 2003-2016 addressing various aspects of economic load dispatch using the PSO algorithm have been presented in this paper along with a brief discussion of a simple ELD model and the classical PSO algorithm. ELD problems have been identified and classified into five important groups. These groups are (i) single objective economic dispatch, (ii) dynamic economic load dispatch, (iii) economic dispatch with non-conventional sources, (iv) multi-objective environmental/economic dispatch and (v) economic dispatch of micro-grids. An attempt has been made to include more and more descriptions to point out the unique and important aspects of each paper considered. In summary, there are several promising approaches with the help of PSO in smart grid technologies for further progress. Some of the areas are as follows:
	
	\begin{enumerate}
		\item [(a)]
		\textbf{Electric Vehicle Charging in Smart Grids:} PSO can be applied to charging optimization of electric vehicle (charging plan of each vehicle while satisfying the requirements of the individual vehicle owners without distribution network congestion) and its coordination to minimize power losses and improve the voltage profile of the smart grid.
		\item [(b)]
		\textbf{Protection of Smart Grid:} Since smart grids are more flexible, fault current levels in the grid are variable. The conventional protection system may not be effective in such a situation. Research can be focused to apply improved PSO algorithms to limit the fault currents in smart grids by the size of thyristor-controlled impedance
		\item [(c)]
		\textbf{Resource Scheduling:} Coordinated scheduling of distributed energy resources (including residential energy sources) and balancing of supply and demand across time with the help of optimization algorithms.
	\end{enumerate}
	
	\bibliography{review_ref}
	\bibliographystyle{IEEEtran}
\end{document}